\documentclass[letterpaper]{article}

\usepackage{aaai2026}
\usepackage{times}
\usepackage{helvet}
\usepackage{courier}  %新加
\usepackage{amsmath}  % 新加
\usepackage{amssymb}  % 新加

\usepackage{algorithm}
\usepackage{algpseudocode}

\usepackage[utf8]{inputenc}
\usepackage{verbatim}

\usepackage{booktabs}

\usepackage{multirow}
\usepackage{makecell}

\usepackage[hyphens]{url}  % DO NOT CHANGE THIS
\usepackage{graphicx} % DO NOT CHANGE THIS
\usepackage{caption} % DO NOT CHANGE THIS AND DO NOT ADD ANY OPTIONS TO IT

\usepackage{natbib}

\usepackage{hyperref}

\nocopyright

\begin{document}
\title{Task-Aware LLM Council with Adaptive Decision Pathways for Decision Support}
\author {
    % Authors
    Wei Zhu\textsuperscript{\rm 1,2},
    Lixing Yu\textsuperscript{\rm 1,2},
    Hao-Ren Yao\textsuperscript{\rm 3},
    Zhiwen Tang\textsuperscript{\rm 1,2}\thanks{Corresponding author.},
    Kun Yue\textsuperscript{\rm 1,2}
}
\affiliations {
    % Affiliations
    \textsuperscript{\rm 1}School of Information Science and Engineering, Yunnan University, Kunming, China\\
    \textsuperscript{\rm 2}Yunnan Key Laboratory of Intelligent Systems and Computing, Kunming, China\\
    \textsuperscript{\rm 3}Language Technologies Institute, Carnegie Mellon University, Pittsburgh, PA, USA\\
    % Email or code link (optional but recommended in this block)
}

\maketitle
\begin{abstract}

 language models (LLMs) have shown strong capabilities across diverse decision-making tasks. However, existing approaches often overlook the specialization differences among available models, treating all LLMs as uniformly applicable regardless of task characteristics. This limits their ability to adapt to varying reasoning demands and task complexities. In this work, we propose Task-Aware LLM Council (TALC), a task-adaptive decision framework that integrates a council of LLMs with Monte Carlo Tree Search (MCTS) to enable dynamic expert selection and efficient multi-step planning. Each LLM is equipped with a structured success memory profile derived from prior task trajectories, enabling semantic matching between current reasoning context and past successes. At each decision point, TALC routes control to the most contextually appropriate model and estimates node value using a dual-signal mechanism that fuses model-based evaluations with historical utility scores. These signals are adaptively weighted based on intra-node variance and used to guide MCTS selection, allowing the system to balance exploration depth with planning confidence. Experiments on WebShop, HumanEval, and the Game of 24 demonstrate that TALC achieves superior task success rates and improved search efficiency compared to strong baselines, validating the benefits of specialization-aware routing and adaptive planning.

\end{abstract}

\section{Introduction}
Large language models (LLMs) have shown remarkable capabilities in a variety of decision-making tasks, including program synthesis, symbolic reasoning, and goal-directed planning \citep{wei2022chaincot, kojima2022zeroshot-cot, ning2023skeleton-cot, yao2023react, shinn2023reflexion}. Their ability to reason over complex inputs and generate coherent action sequences has made them attractive components in autonomous agents. To support long-horizon tasks, recent work \citep{cot-sc2023,yao2023tree, hao2023reasoning, gan2025master, shi2025monte-ICLR-MCTS} has explored integrating LLMs with structured planning algorithms such as Monte Carlo Tree Search (MCTS), enabling decision paths to be generated incrementally through interaction with the environment. As the demand for general-purpose decision agents increases, developing principled methods that can leverage the reasoning capabilities of LLMs in a structured and scalable way becomes a central research challenge.

Despite this growing interest, most existing approaches \citep{yao2023tree, hao2023reasoning, zhou2024language, gan2025master} treat LLMs as monolithic agents and apply them uniformly across tasks, overlooking the nuanced performance differences that exist even among general-purpose models. In practice, different LLMs exhibit varying degrees of effectiveness depending on the task domain and reasoning requirements. Some LLMs are more proficient at symbolic manipulation, some others are better at language generation or procedural synthesis. Failing to account for these distinctions leads to brittle performance: a model that excels on one task may perform poorly on another when applied indiscriminately. Furthermore, these systems typically follow static inference pipelines, invoking the same model with a fixed reasoning depth regardless of task complexity. This rigid setup ignores two key sources of variability in decision tasks: (1) the suitability of different models for different reasoning subtasks, and (2) the variation in how much planning is needed to reach a satisfactory outcome. As a result, current LLM-based agents often suffer from inefficient search behavior, error accumulation from early-stage missteps, and unnecessary computational overhead on simple tasks. Addressing these limitations calls for a more adaptive approach that can dynamically leverage model-specific strengths and modulate reasoning depth based on contextual cues.

To address these limitations, we propose a task-adaptive decision framework built around a council of LLMs, each treated as a profiled expert with empirically grounded strengths. Instead of assuming uniform capabilities, we construct a structured success memory for each model, composed of trajectory segments from previously successful task completions. These segments serve as contextual exemplars that encode the kinds of reasoning patterns where each model has demonstrated competence. At inference time, the current task context is semantically matched against all expert profiles, and the most contextually aligned model is selected to act. This dynamic model routing forms the core of our framework, enabling fine-grained task-aware delegation that reflects the functional strengths of each LLM.

To support multi-step reasoning, we couple expert routing with adaptive planning guided by dual value signals—real-time expert evaluation and memory-based priors. These signals jointly inform node selection, enabling the planner to dynamically adjust its search depth based on intermediate reasoning quality. This allows the system to focus on high-potential trajectories without exhaustive exploration, improving both decision quality and computational efficiency.

We evaluate our framework on three challenging domains: WebShop, HumanEval, and the Game of 24. These domains vary significantly in modality, reasoning style, and task structure, providing a rigorous testbed for adaptive decision-making. Across all benchmarks, our approach achieves consistently higher task success rates and improved search efficiency compared to strong baselines that rely on single-model inference or static routing policies. These results highlight the benefit of leveraging model-specific expertise through structured profiling and demonstrate that context-aware model selection, when combined with adaptive planning, offers a promising direction for building more capable and efficient LLM-based agents.

% Our contributions are summarized as follows:

% \begin{itemize}
%     \item We propose a unified framework that integrates model specialization and planning adaptation to improve decision-making with large language models. The framework  leverages empirical model strengths and adapt to varying task complexities, enabling more accurate and efficient reasoning.
%     \item  We construct a structured LLM council with success-based profiling, where each model maintains a memory of successful reasoning segments. At each decision point, we perform semantic retrieval against these profiles to dynamically route control to the most contextually competent expert.
%     \item We develop a dual-signal value estimation mechanism for adaptive search control, which fuses real-time expert feedback with historical utility priors to guide MCTS. This enables the system to modulate planning depth based on task-specific cues and intermediate progress.
%     \item We validate our approach across three diverse benchmarks, demonstrating consistent improvements in task success rate and planning efficiency over strong single-model and static-routing baselines.
% \end{itemize}

\section{Related Work}

\subsection{LLM as a Single-Agent System}
Initial efforts on LLM-based reasoning aimed to improve correctness via in-context exemplars \citep{brown2020lfewshot} and step-by-step rationales \citep{wei2022chaincot,kojima2022zeroshot-cot}. To enhance adaptability, methods such as ReAct \citep{yao2023react} and Reflexion \citep{shinn2023reflexion} introduced interaction and self-correction. More structured reasoning emerged with AgentKit \citep{wu2024agentkit}, which enables dynamic DAG-based inference over natural language subgoals, and Tree-of-Thoughts (ToT) \citep{yao2023tree}, which organizes generation into a search tree with backtracking. 

Recent work applies Monte Carlo Tree Search (MCTS) to further systematize reasoning: RAP \citep{hao2023reasoning} treat the LLM as a world model and use reward signals to guide search, LATS \citep{zhou2024language} integrate planning and reflection via learned value estimates, and MC-DML \citep{shi2025monte-ICLR-MCTS} enhance exploration with memory-augmented MCTS in text-based games. 

These approaches, while effective, rely on a single-agent paradigm. In contrast, our method introduces a council of specialized experts, enabling broader task coverage and more flexible reasoning coordination.

\subsection{Multi-Agent Collaboration}
Multi-agent frameworks enhance reasoning diversity, adaptability, and robustness by coordinating multiple LLMs or specialized modules. Early systems such as ChatDev \citep{qian2023chatdev} and MetaGPT \citep{hong2023metagpt} adopt fixed-role designs, structuring workflows into predefined stages (e.g., planning, coding, testing). While effective in constrained domains, they lack the flexibility needed for dynamic, open-ended tasks.

Recent advances emphasize adaptive coordination. AutoGen \citep{wu2023autogen} enables runtime role negotiation and task delegation. MAd \citep{li2024MAD} introduces adversarial debate to expose reasoning flaws. MoA \citep{wang2024mixture—moa} presents a hierarchical Mixture-of-Agents framework, where agents at each layer generate responses based on aggregated outputs from the previous layer—capturing emergent collective intelligence. MASTER \citep{gan2025master} augments MCTS with reward-sensitive multi-agent selection. 

Unlike these approaches that rely on static role assignments or costly coordination, our method introduces task-aware expert routing to dynamically select the most suitable models, enabling lightweight and flexible heterogeneous collaboration.

\subsection{Memory-Augmented LLMs}
Self-adaptive agents leverage autonomous memory management to improve reasoning over time through experience collection, example adaptation, and selective curation. ExpeL \citep{zhao2024expel} enables agents to accumulate success and failure trajectories in an experience pool, perform natural language operations (e.g., add, upvote, edit) to refine knowledge, and retrieve task-relevant memories during inference—all without model parameter updates. Many-shot in-context learning techniques \citep{agarwal2024manyshot} further enhance memory utility by scaling exemplars (up to thousands of shots) and incorporating model-generated rationales, boosting adaptability with limited human data. Traj-Bootstrap \citep{sarukkai2025-Self-generated} improves memory curation via population-based optimization and exemplar filtering, dynamically retrieving relevant trajectories to guide sequential decisions. AZR \citep{zhao2025absolute} advances autonomy by enabling self-generated task creation, self-play memory building, and verifiable feedback loops without external data dependence.

Despite these advances, many memory-augmented methods rely on extensive offline memory collection and require maintaining large, complex memory stores, incurring significant engineering costs. In contrast, our approach collects lightweight successful trajectories  online during inference, forming expert profiles that efficiently guide routing and decision making without offline preparation.

\section{Preliminaries}

\begin{figure*}[t]  % t表示放在页面顶端
    \centering
    \includegraphics[width=0.8\textwidth]{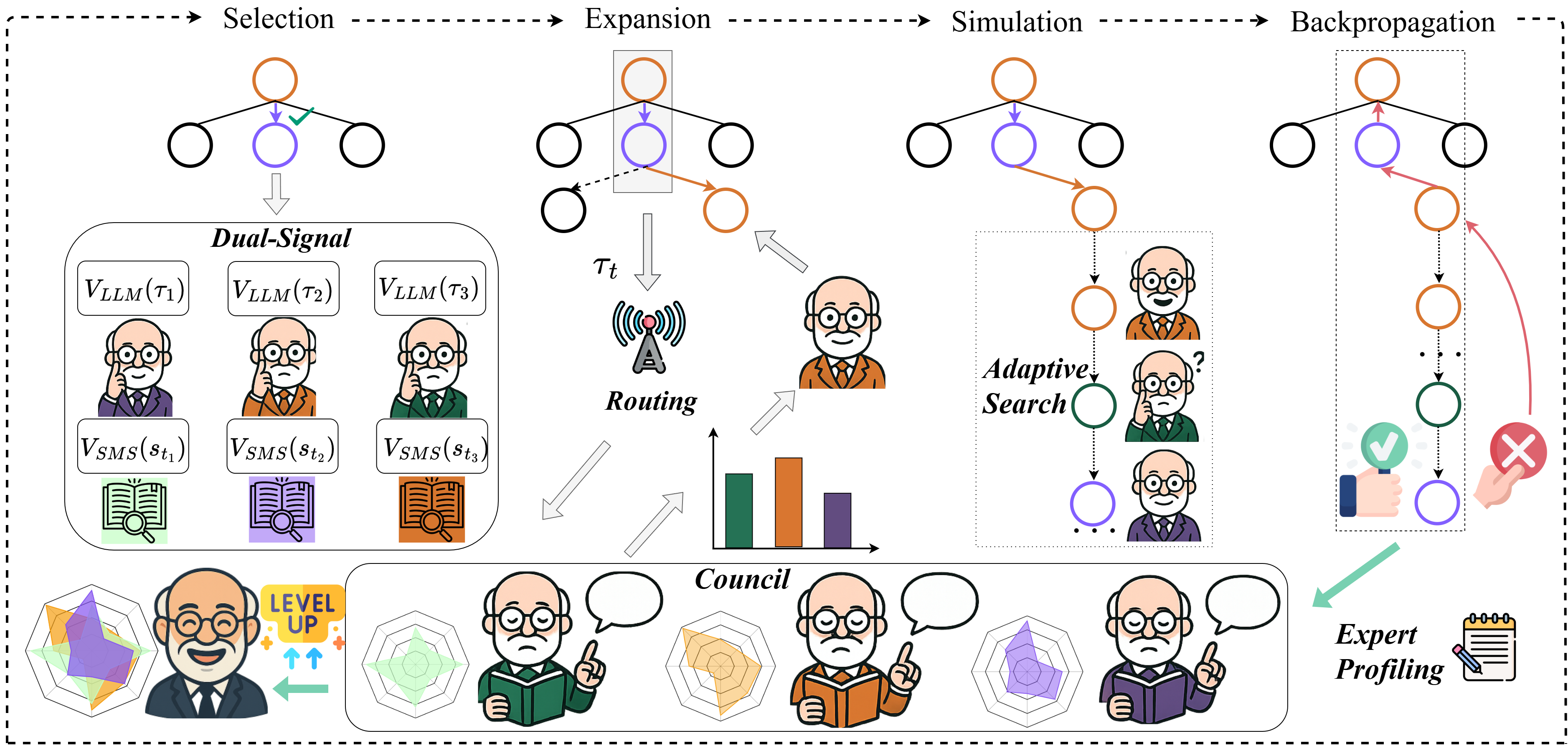}  % 替换为实际图片文件名
    \caption{Overview of TALC.}
    \label{fig:overview}
\end{figure*}

\subsection{Reinforcement Learning and LLM Agents}
Sequential decision-making is commonly formalized as an agent–environment interaction over discrete time steps. At each step \( t \), the agent observes \( o_t \in \mathcal{O} \), takes action \( a_t \in \mathcal{A} \), receives reward \( r_t = R(o_t, a_t) \), and transitions to \( o_{t+1} \sim \mathcal{T}(\cdot \mid o_t, a_t) \), with \( \mathcal{T} \) denoting transition dynamics.

The agent follows a policy \( \pi(a \mid o) \) and aims to maximize the expected return:
\begin{equation}
    \mathbb{E}_{\pi} \left[ \sum_{t=0}^{\infty} \gamma^t R(o_t, a_t) \right], \quad \gamma \in [0,1)
\end{equation}

Large language models (LLMs) have been widely used as agents in such settings, where observations are textual inputs and actions are generated via autoregressive decoding:
\begin{equation}
a_t \sim \pi_\theta(\cdot \mid o_t)
\end{equation}
where \( \pi_\theta \) is the model-induced policy parameterized by \(\theta\). This formulation enables interaction with text-based environments without explicit state representations or environment models.

While LLM agents offer strong contextual reasoning, they lack planning, memory, and exploration mechanisms, which highlights the need for structured decision-time search, introduced next.

\subsection{Monte Carlo Tree Search (MCTS)}
\label{mcts}
Monte Carlo Tree Search (MCTS) \citep{coulom2006MCTS} is a sample-based planning algorithm widely applied in complex sequential decision-making domains such as Go \citep{prelimi:go} and Atari games \citep{prelimi:Atari}. MCTS incrementally constructs a search tree, where each node represents a state and edges correspond to actions.

Each MCTS iteration simulates a trajectory from the root node through the following four phases:

(1) \textbf{Selection:} Traverse the current tree from the root by selecting child nodes that maximize a selection criterion, typically the Upper Confidence Bound for Trees (UCT) \citep{prelimi:UCT}.
% \begin{equation}\label{eq:uct}
% \text{UCT}(s) = V(s) + c \cdot \sqrt{\frac{\log N(p)}{N(s)}}
% \end{equation}
% where $V(s)$ is the current value estimate, $N(s)$ is the visit count of node $s$, $N(p)$ its parent's visit count, and $c$ is an exploration constant.

(2) \textbf{Expansion:} When reaching a leaf node that is not terminal, expand the tree by adding new child nodes through sampling actions;

(3) \textbf{Simulation (Rollout):} From the expanded node, the algorithm simulates a trajectory by selecting actions without growing the tree until termination, using the outcome to estimate the node’s value.

(4) \textbf{Backpropagation:} Propagate the obtained return backward along the visited path to update value estimates $Q(s)$
and visit counts $N(s)$ at each node.
After each simulation, the value function is updated via incremental averaging:
\begin{equation}\label{eq:back}N(s) \leftarrow N(s) + 1, \quad 
Q(s) \leftarrow \frac{(N(s) - 1) \cdot Q(s) + r}{N(s)}
\end{equation}
where $r$ is the return obtained from the simulated trajectory.

Conventional MCTS requires an environment model to simulate and reset states. In contrast, language-based environments represent state via interaction histories that can be deterministically reconstructed by replaying prior text. This unique property enables effective decision-time planning with language models without an explicit transition model.

\section{Methodology}

We propose a decision-making framework that improves planning efficiency by dynamically selecting the most competent LLM for each decision step and adaptively controlling the search trajectory. The core of our design is a council of LLM agents, each maintaining an expert profile distilled from successful decision traces. These profiles support task-aware model routing by matching current reasoning context to past exemplars, while downstream planning is guided by a dual-signal value estimator that fuses model feedback with memory-based priors. These components enable fine-grained delegation and depth-aware search, forming the foundation of our method.

\subsection{Expert Profiling for LLM council}  \label{sec:expert_profiles}

To enable dynamic model selection grounded in prior empirical evidence, we organize multiple general-purpose LLMs into a collaborative council. Each member $M_j$  is associated with an expert profile $\mathcal{P}_j$, which serves as a compact memory of the model's past decision-making successes. Rather than relying on predefined task labels or static capabilities, our framework learns each expert’s strength through structured traces distilled from successful episodes.
 
These traces are represented as Success Memory Segments (SMS), which are partial trajectories extracted from rollouts that culminated in correct or satisfactory outcomes. Given a successful trajectory of depth $d$, we decompose it into prefix-aligned segments:
\begin{equation}\label{eq:tau_defn}
\tau_t=\{(o_0,a_0),(o_1,a_1),\ldots,(o_{t-1},a_{t-1})\}
\end{equation}
where $t=1,2, \ldots, d$. Here, $o_i$ and $a_i$ denote the observation and the corresponding action at step $i$. Each prefix $\tau_t$ reflects the incremental reasoning context up to depth $t$. These SMS units serve as semantically grounded representations of reasoning progress, enabling fine-grained generalization even across tasks with dissimilar surface structures.

To assess the utility of each segment, we track its retrieval history and assign a weighted score based on how often and how effectively it contributes to task success:
\begin{equation} \label{eq:sms_utility}
V_{\text{SMS}}(\tau) = \frac{\sum_{i \in \mathcal{U}(\tau)} y_i \cdot u_i(\tau)}{\sum_{i \in \mathcal{U}(\tau)} u_i(\tau)}
\end{equation}

In this formulation, $\mathcal{U}(\tau)$ is the set of prior episodes where $\tau$ was retrieved, $y_i \in {0, 1}$ indicates whether episode $i$ led to task success, and $u_i(\tau)$ is the number of times $\tau$ was matched or reused during episode $i$. This metric favors trajectory segments that are not only frequently recalled but also predictive of successful outcomes.

% To ensure efficiency and memory precision, we retain only the top-$k$ scoring segments for each model. The resulting expert profile $\mathcal{P}_j$ for model $M_j$ consists of a curated subset of high-quality SMS, serving as an empirical prior over model competence. These profiles support both task-aware model routing (Section 4.2) and value-based planning (Section 4.3) by grounding decisions in past successful experience.

To ensure both efficiency and memory precision, each member’s expert profile $\mathcal{P}_j$ is maintained in a vector database that stores SMS in real time. The framework allows dynamic control over memory size via $V_{\text{SMS}}(\tau)$-guided pruning, enabling flexible adaptation to task demands. These profiles support task-aware model routing (Section 4.2) and value-based planning (Section 4.3), grounding decisions in accumulated successful experience.

\subsection{Task-Aware Council Routing}

Building on the expert profiles, we introduce a retrieval-augmented routing mechanism that selects the most suitable model at each decision step. Although the models are not explicitly specialized, they often demonstrate distinct empirical strengths. Our approach dynamically delegates decision authority by aligning the current reasoning context with trajectory segments that have previously led to successful outcomes.

At each decision point $s_t$, the system queries the council $\mathcal{M}=\{M_1, \cdots , M_n\}$ using the current trajectory prefix $\tau_t$ (Eq. \ref{eq:tau_defn}). Each expert $M_j$ compares this prefix against its success profile $\mathcal{P}_j$ , which consists of previous trajectory segments from successful completions. To determine its relevance, the expert identifies the best-matching exemplar within its profile:
\begin{equation}
\mu_j = \max_{\tau_j^{(i)} \in \mathcal{P}j} \text{sim}(\tau_t, \tau_j^{(i)})
\end{equation}
These maximal similarity scores $\{\mu_j\}_{j=1}^n $are then normalized into a soft routing distribution across all council members:
\begin{equation}
P(M_j \mid \tau_t) = \frac{\exp(\mu_j / T)}{\sum_{k=1}^n \exp(\mu_k / T)}
\end{equation}
Here, $\text{sim}(\cdot, \cdot)$denotes a semantic similarity function that compares partial trajectories, and $T>0$ is a temperature parameter that controls the confidence level of expert selection. This formulation ensures that the most semantically aligned expert receives a higher routing probability, enabling task-sensitive delegation grounded in historical success patterns.

Based on the soft routing distribution $P(M_j | \tau_t)$, we sample an expert $M_* \sim P(M_j | \tau_t)$ to act at the current step. From the selected expert’s profile $\mathcal{P}_*$, we retrieve the most relevant success memory segment:
\begin{equation}
\tau^* = \arg\max_{\tau^{(i)} \in \mathcal{P}_{*}} \text{sim}(\tau_t, \tau^{(i)})
\end{equation}

% This exemplar $\tau^*$ serves as an informative reference and is injected into the prompt using a structured composition function:
% \begin{equation}
% \text{Prompt}_t = \Phi(\tau_t, \tau^*)
% \end{equation}
% where $\Phi$ denotes a formatting operation that fuses the current decision trace with the retrieved exemplar while maintaining logical separation and semantic coherence.

The retrieved exemplar $\tau^*$ is incorporated into the prompt as a contextual reference to guide next-step reasoning. Rather than treating it as a sequential continuation, we encode it as a structurally separated memory trace that reflects a previous successful trajectory. This composition is inspired by memory replay mechanisms in reinforcement learning, but adapted to the inference-time setting: $\tau^*$ is selected for its contextual relevance, embedded alongside the current trajectory $\tau_t$ in role-specific regions of the prompt, and annotated with minimal instructions to preserve semantic independence. This setup allows the model to draw inductive cues from prior successes without conflating past and present reasoning paths, which effectively supports retrieval-augmented generalization in multi-step decision-making.

By transforming expert selection into a retrieval-augmented routing process, this mechanism ensures that each decision step benefits from both task-specific priors and historical success patterns. The LLM council thus functions not only as a set of candidate decoders but as a context-sensitive planner, dynamically assigning decision authority to the most capable model at each point in the trajectory.

\subsection{Dual-Signal Value Estimation} \label{sec:dual-signal}
While task-aware model selection enables the system to choose competent LLMs at each decision point, it does not by itself guarantee effective long-term planning. In complex decision tasks that require multi-step reasoning, it is crucial to explore potential future trajectories and evaluate their expected utility before committing to a particular action. To this end, we incorporate MCTS as a planning backbone, allowing the system to simulate and compare possible decision pathways. MCTS provides a principled framework for balancing exploration and exploitation during search, but its effectiveness hinges on the ability to accurately estimate the value of intermediate states. To support adaptive planning depth and informed model invocation, we design a dual signal value estimation mechanism that combines model-level feedback and historical task-specific success priors. This hybrid approach enables the search process to dynamically prioritize promising branches, thereby improving efficiency and reducing unnecessary exploration.

In each node expansion step, the system generates a set of candidate child nodes \( \mathcal{C}(s_{t-1}) = \{s_t^{(1)}, s_t^{(2)}, \ldots, s_t^{(K)} \} \), corresponding to different possible actions taken by the parent node \( s_{t-1} \). For each candidate node \( s_t \in \mathcal{C}(s_{t-1}) \), we compute two complementary value signals to reflect both structural coherence and empirical task utility:

\begin{itemize}
\item \textbf{LLM Evaluation Score}. \( V_{\text{LLM}}(s_t) \) is obtained by random sampling an expert \( M_j \) from the council and prompting it to assess the plausibility of the trajectory leading to \( s_t \). This value reflects the expert model’s judgment on the semantic validity and structural consistency of the current decision prefix \( \tau_t \). Random sampling encourages diverse assessments and mitigates over-reliance on a single expert.

\item \textbf{SMS-Based Utility Score}. \( V_{\text{SMS}}(\tau_t) \) is computed by querying the success memory of the currently routed expert using the current trajectory prefix \( \tau_t \). The top-matched prefix $\tau^*$ provides a historical utility signal reflecting the expert’s prior success under similar contexts, as defined in Eq.~\ref{eq:sms_utility}.
\end{itemize}

While each value signal provides useful guidance, their relative discriminative power may vary across tasks and search stages. To adaptively balance their contributions, we adopt a variance-based fusion mechanism that adjusts weights according to signal variability among sibling nodes. We compute their standard deviations over the current candidate set \( \mathcal{C}(s_{t-1}) \), denoted \( \sigma_{\text{LLM}} \) and \( \sigma_{\text{SMS}} \) respectively. These are used to derive a fusion weight \( \alpha =  \frac{\sigma_{\text{LLM}}}{\sigma_{\text{LLM}} + \sigma_{\text{SMS}}}\in [0,1] \), which adaptively balances the two signals. 

To ensure comparability, we normalize both \( V_{\text{LLM}}(s_t) \) and \( V_{\text{SMS}}(\tau_t) \) over the same set \( \mathcal{C}(s_{t-1}) \), denoted as \( \widetilde{V}_{\text{LLM}}(s_t) \) and \( \widetilde{V}_{\text{SMS}}(\tau_t) \). The final fused value for each child node is given by:
\begin{equation}
    Q(s_t) = \alpha \cdot \widetilde{V}_{\text{LLM}}(s_t) + (1 - \alpha) \cdot \widetilde{V}_{\text{SMS}}(\tau_t)
\end{equation}

This formulation ensures that the signal with greater discriminative power, as quantified by its variance across sibling nodes, receives proportionally higher influence in the fusion process. The resulting value $Q(s_t)$ is subsequently employed within the UCT formula (Eq.~\ref{eq:uct}) to guide node selection in MCTS.

By grounding value estimation in both immediate expert feedback and long-term empirical evidence, our dual-signal mechanism ensures that the search process remains both responsive to local decision quality and aligned with global task success trends. This, in turn, supports efficient and robust planning behavior in tasks with high structural variability.

\begin{figure*}[t]  % t表示放在页面顶端
    \centering
    \includegraphics[width=0.9\textwidth]{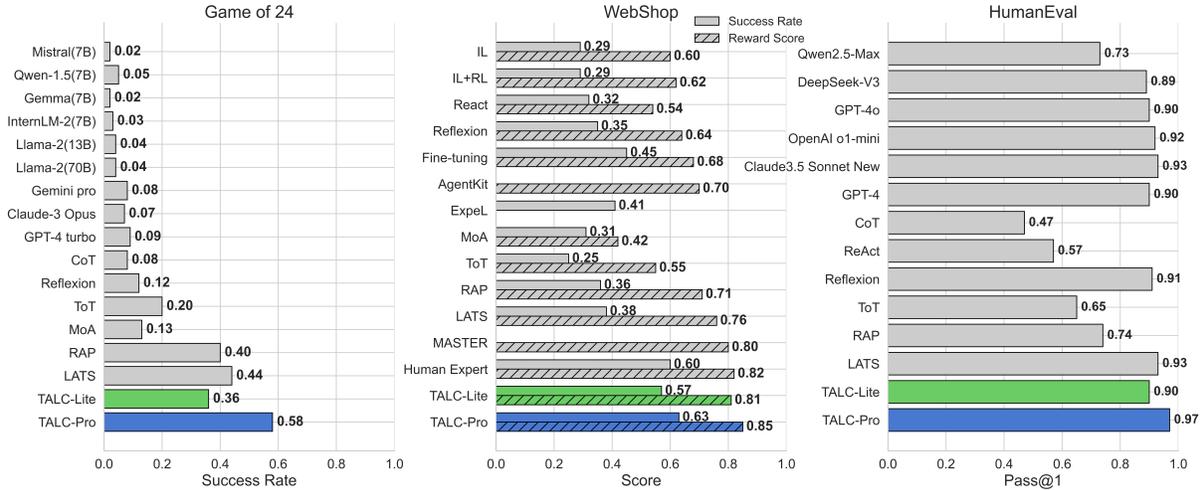}  % 替换为实际图片文件名

    \caption{Main Results. Metrics include success rate (Game of 24 and WebShop) and pass@1 (HumanEval). The WebShop reward score ($r \in [0,1]$) reflects partial fulfillment of user-specified attributes. Some baselines (e.g., MASTER) report only a subset of metrics due to unavailable code and are shown accordingly.}
    \label{fig:main_results}
\end{figure*}

\subsection{Adaptive Search via Controlled Expansion}

% A central challenge in planning-based decision-making is determining how far the search should proceed. In many real-world tasks, the appropriate depth of reasoning is highly context-dependent. Simple problems may require only a few steps to reach a satisfactory solution, while more complex ones demand deeper exploration. Traditional MCTS-based systems often rely on fixed or manually tuned search depths, which can result in either premature termination or unnecessary computational overhead. In contrast, our framework enables \emph{adaptive search depth} by improving the quality of local decisions through expert-guided routing and value estimation.

% While the branching factor in our MCTS implementation remains constant across nodes, the actual depth of the search tree is determined dynamically during planning. Specifically, as described in Section~4.3, each candidate node is assigned a value based on a fusion of expert model evaluation and empirical utility. These values influence node selection through the UCT algorithm:

A fundamental challenge in planning-based decision-making is how to determine the appropriate depth of search. In practice, task complexity varies widely. Some problems can be resolved with only a few reasoning steps, while others require extended exploration to reach satisfactory solutions. Traditional MCTS approaches often rely on fixed-depth heuristics, which may cause premature termination on complex tasks or unnecessary computation on simpler ones. Our framework avoids this by enabling adaptive search depth, where the reasoning trajectory evolves according to the quality of intermediate decisions.

This adaptivity is achieved by coupling task-aware model routing with dual-signal value estimation. Each candidate node receives a value $Q(s)$ that combines the expert's model-based judgment and historical trajectory utility, which then feeds into the UCT selection rule:
\begin{equation}
\text{UCT}(s) = Q(s) + c \cdot \sqrt{\frac{\log N_p}{N_s}}
\label{eq:uct}
\end{equation}

Paths with consistently high values, which are enabled by accurate model invocation and effective memory reuse, are prioritized early, leading to concentrated exploration and shallow, efficient trees. In contrast, when uncertainty arises due to weak model-task alignment or ambiguous observations, lower value estimates push the planner toward deeper, more exploratory search.

In this way, the depth of search is not predetermined. It emerges from the interaction between model competence, value estimation, and trajectory feedback. The system learns to allocate computational effort based on task difficulty, converging quickly when early decisions are sound and extending deeper only when necessary. This controlled expansion mechanism improves planning efficiency while preserving flexibility across a wide range of decision tasks.

\section{Experiments}

\subsection{Experiment Settings}
We follow the experimental setup adopted in prior baselines \citep{yao2023react, shinn2023reflexion,yao2023tree,zhou2024language} to enable fair and consistent comparison. Specifically, we evaluate our approach on three text-based reasoning benchmarks with distinct characteristics: \textbf{Game of 24} \citep{yao2023tree}, a symbolic arithmetic task where four given numbers must be combined using basic operations to produce 24; \textbf{WebShop} \citep{yao2022webshop}, an interactive decision-making task in a simulated online shopping environment; and \textbf{HumanEval} \citep{chen2021humaneval}, a code generation benchmark measuring functional correctness. 

% Performance is reported using task-specific success metrics: exact-match success rate for Game of 24, the proportion of successful attribute-matching purchases for WebShop, and pass\@1 for HumanEval. For WebShop, we additionally report a normalized reward score $ r \in \left [0,1 \right ]$, which reflects partial fulfillment of user-specified attributes. To mitigate the effects of stochastic generation, each experiment is repeated three times on the same data splits, and mean scores are reported.

% To assess robustness under different computational regimes, we consider two deployment configurations. \textbf{TALC‑Lite} uses open-weight language models deployable on local hardware, including Qwen2.5‑7B‑Instruct‑1M \citep{yang2024qwen2-5-1M}, Mistral‑7B‑Instruct‑v0.3 \citep{jiang2023mistral7b}, and Llama‑3.1‑8B‑Instruct \cite{grattafiori2024llama3herdmodels}. This configuration offers fast, cost-effective inference while supporting modular expert composition. In contrast, \textbf{TALC‑Pro} leverages API-based access to high-capacity proprietary models, including GPT‑4 \citep{achiam2023gpt4}, Qwen‑Max (2024‑09‑19) \citep{yang2024qwen2-5}, and DeepSeek‑V3 (2025‑03‑24) \citep{liu2024deepseek}, enabling stronger reasoning and language understanding. All experiments share the same agent routing and search configurations to ensure comparability across setups. To mitigate the effects of stochastic generation, each experiment is repeated three times on the same data splits, and mean scores are reported.

To evaluate robustness across computational settings, we compare two variants. \textbf{TALC‑Lite} uses locally deployable open-weight models, comprising of Qwen2.5‑7B‑Instruct‑1M~\citep{yang2024qwen2-5-1M}, Mistral‑7B‑Instruct‑v0.3~\citep{jiang2023mistral7b}, and Llama‑3.1‑8B‑Instruct~\citep{grattafiori2024llama3herdmodels}, enabling efficient and modular inference. \textbf{TALC‑Pro} accesses proprietary APIs with GPT‑4~\citep{achiam2023gpt4}, Qwen‑Max~\citep{yang2024qwen2-5}, and DeepSeek‑V3~\citep{liu2024deepseek} for enhanced reasoning. Both use identical routing and planning setups. Results are averaged over three runs to reduce stochasticity. We adopt reasoning-oriented methods and strong single-model methods as our baselines. Details about experimental settings and baselines can be found in Appendix G.

\subsection{Main Results}
% 用图片以增加和原来工作的区分度，但是不能是你现在的这个，主要问题：
% 1、横坐标排序方式难以justify, 
% 2、LATS，RAP等，横坐标位置为什么是图中那样
% 3、绿色虚线含义不明确
% 4、作为主实验，不要强调council大小的影响，否则会和已有工作太类似，可能只需要保留small*3, big，可加入不同routing方式、MoA的点

Figure~\ref{fig:main_results} presents a comparison of TALC‑Lite and TALC‑Pro across three benchmarks.

On \textbf{Game of 24}, TALC‑Pro achieves a new state-of-the-art, surpassing all prior reasoning methods and significantly outperforming powerful monolithic models. This highlights the strength of our search-based coordination over direct generation. On \textbf{WebShop}, TALC‑Pro attains the highest reward (0.85) and success rate (0.63), demonstrating superior long-horizon planning. TALC‑Lite also outperforms competitive methods like LATS and MASTER, approaching human expert performance under stricter compute constraints. On \textbf{HumanEval}, TALC‑Pro reaches a pass@1 of 0.97, outperforming all reasoning baselines and even large proprietary models. Remarkably, TALC‑Lite matches GPT‑4o, confirming the efficiency and competitiveness of our lightweight variant.

Overall, these results demonstrate TALC’s strong generalization across different domains and consistently high performance under both configurations.

\subsection{Ablation Study}
% 从两个维度，

% 一个维度证明Council有用
% 不同Council 大小
% 用完整Council，但是使用不同Routing方式
% 用完整Council，使用我们提出的Routing方式，但是成功轨迹注入方法不同

% 另一个维度证明双信号MCTS有用
% 用Council但是不使用MCTS（MoA）
% 用Council，用MCTS，但是不使用双信号

% 以上实验是否三个数据集都有？有的话就都放上来，用表格的形式
% 两个维度可以考虑拆成两个表
% ------------------------------

% \begin{table}[t]
% \centering
% \small
% \caption{Ablation on the Expert Council mechanism.}
% \label{tab:ablation_counsil}
% \begin{tabular}{lccc}
% \toprule
% Configuration & Game of 24 & WebShop & HumanEval \\
% \midrule
% Single (Qwen) & 0.09 & 0.33 & 0.66 \\
% Single (Mistral) & 0.05 & 0.30 & 0.53 \\
% Single (Llama) & 0.08 & 0.02 & 0.52 \\
% Dual (Qwen+Mistral) & 0.19 & 0.51 & 0.83 \\
% Dual (Qwen+Llama) & 0.24 & 0.48 & 0.80 \\
% Dual (Mistral+Llama) & 0.13 & 0.43 & 0.75 \\
% \midrule
% Random & 0.31 & 0.48 & 0.85 \\
% Round-Robin & 0.29 & 0.43 & 0.81 \\
% Collaborative & 0.21 & 0.39 & 0.72\\
% Voting & 0.32 & 0.46 & 0.73 \\
% \midrule
% No Memory Replay & 0.34 & 0.51 & 0.80 \\
% Success+Failure & 0.25 & 0.28 & 0.70 \\
% Only Failure & 0.26 & 0.31 & 0.77 \\
% \midrule
% \textbf{Full} & \textbf{0.36} & \textbf{0.57} & \textbf{0.90}\\
% \bottomrule
% \end{tabular}
% \end{table}

\begin{table}[t]
\centering
\small
\setlength{\tabcolsep}{2pt}
\begin{tabular}{p{1.5cm}lccc}
\toprule
\textbf{Module} & \textbf{Configuration} & \textbf{Game of 24} & \textbf{WebShop} & \textbf{HumanEval} \\
\midrule
% 用\parbox实现Council Size换行，避免格式冲突
\multirow{6}{*}[3ex]{\parbox{1.3cm}{\centering \textbf{Council \\ Size}}} & 
Single (Q) & 0.09 & 0.33 & 0.66 \\
& Single (M) & 0.05 & 0.30 & 0.53 \\
& Single (L) & 0.08 & 0.02 & 0.52 \\
& Dual (Q+M) & 0.19 & 0.51 & 0.83 \\
& Dual (Q+L) & 0.24 & 0.48 & 0.80 \\
& Dual (M+L) & 0.13 & 0.43 & 0.75 \\
\midrule
\multirow{4}{*}[2ex]{\parbox{1.3cm}{\centering \textbf{Routing \\ Strategy}}} & 
Random & 0.31 & 0.48 & 0.85 \\
& Round-Robin & 0.29 & 0.43 & 0.81 \\
& Collaborative & 0.21 & 0.39 & 0.72 \\
& Voting & 0.32 & 0.46 & 0.73 \\
\midrule
\multirow{3}{*}[1ex]{\parbox{1.3cm}{\centering \textbf{Memory \\ Utilization }}} & 
No use  & 0.34 & 0.51 & 0.80 \\
& S+F & 0.25 & 0.28 & 0.70 \\
& F Only & 0.26 & 0.31 & 0.77 \\
\midrule
& \textbf{Full} & \textbf{0.36} & \textbf{0.57} & \textbf{0.90} \\
\bottomrule
\end{tabular}
\caption{\centering Ablation on the Expert Council mechanism.}
\small Q: Qwen2.5-7B; M: Mistral-7B; L: Llama3.1-8B; S: Success Memory; F: Failure Memory.
\label{tab:ablation_counsil}
\end{table}

% ---------------------
\begin{table}[t]
\centering
\small
\begin{tabular}{lccc}
\toprule
\textbf{Configuration} & \textbf{Game of 24} & \textbf{WebShop} & \textbf{HumanEval} \\
\midrule
Linear Reasoning & 0.02 & 0.23 & 0.47 \\
\hline
Env & 0.18 & 0.38 & 0.75 \\
LLM-Only & 0.27 & 0.51 & 0.82 \\
SMS-Only & 0.23 & 0.39 & 0.81 \\
\hline
\textbf{Full} & \textbf{0.36} & \textbf{0.57} & \textbf{0.90} \\
\bottomrule
\end{tabular}
\caption{Ablation on the Dual-Signal MCTS component.}
\label{tab:ablation_mcts}
\end{table}

% -------------------------------------
% \begin{table*}[ht]
% \centering
% \begin{tabular}{p{0.15\textwidth} p{0.28\textwidth} p{0.28\textwidth} p{0.28\textwidth}}  
% \toprule
% \multicolumn{4}{c}{
% \parbox{\textwidth}{
% \textbf{Instruction:} I'm looking for hair treatments that are sulfate and paraben free and are of high quality too. I need it in bottle form with 60 capsules, and price lower than \$70.00.
% }
% }\\
% \midrule

% \textbf{1. Query} & 
% search[sulfate and paraben free hair treatments 60 capsules bottle price \textless 70.00.] & 
% search[hair treatments sulfate paraben free 60 capsules bottle high quality price \textless 70.00] & 
% search[hair treatment sulfate free paraben free high quality 60 capsules] \\

% \midrule

% \textbf{2. Filter} & 
% click[Next \textgreater] & 
% click[B07MTPW3Z7] & 
% click[Back to Search] \\

% \midrule

% \textbf{3. Check} & 
% click[B09HWS67T8] & 
% click[1 bottle + 60 capsules] & 
% search[high quality hair treatment sulfate-free paraben-free 60 capsules] \\

% \midrule

% \textbf{4. Final} & 
% click[1 bottle + 60 capsules]... click[Buy Now] & 
% click[Buy Now] & 
% click[Back to Search]... search[hair treatment sulfate-free paraben-free 60 capsules bottle] \\

% \midrule

% \textbf{Result} & 
% \textit{Success} (5 steps, Reward=1.00) & 
% \textit{Partial Match} (Missed quantity, Reward=0.75) & 
% \textit{Inefficient} (Multiple errors, Reward=0.25) \\

% \midrule
%  & \textbf{Ours} & \textbf{MoA} & \textbf{LATS} \\

% \bottomrule
% \end{tabular}

% \caption{Case Study on a WebShop Task}  
% \label{tab:case-study}
% \end{table*}

\begin{table*}[t]  % 双栏用table，table*是跨栏
\centering
\setlength{\tabcolsep}{2pt}  % 压缩列间距
\small  % 缩小整体字体
\begin{tabular}{>{\raggedright\arraybackslash}p{0.1\textwidth}| p{0.22\textwidth} |p{0.25\textwidth} |p{0.28\textwidth}}
\toprule
\multicolumn{4}{c}{
\parbox{0.8\textwidth}{\textbf{Instruction:} I'm looking for hair treatments that are sulfate and paraben free, high quality, bottle form, 60 capsules, price $<$ \$70.00.}  % 简化标题换行
}\\
\midrule
\textbf{1. Query} & 
\textit{search}[sulfate/paraben-free hair treatments, 60 capsules, bottle, price \textless 70] & 
\textit{search}[hair treatments, sulfate paraben-free, 60 capsules, bottle, high quality, price \textless 70] & 
\textit{search}[high-quality, sulfate/paraben-free hair treatment, 60 capsules] \\
%\midrule
\textbf{2. Filter} & 
\textit{click}[Next \textgreater] & 
\textit{click}[B07MTPW3Z7] & 
\textit{click}[Back to Search] \\
%\midrule
\textbf{3. Check} & 
\textit{click}[B09HWS67T8] & 
\textit{click}[1 bottle + 60 capsules] & 
\textit{search}[high-quality, sulfate/paraben-free hair treatment, 60 capsules] \\
%\midrule
\textbf{4. Final} & 
\textit{click}[1 bottle + 60 capsules]... \textit{click}[Buy Now] & 
\textit{click}[Buy Now] & 
\textit{click}[Back to Search]... \textit{search}[sulfate/paraben-free hair treatment, 60 capsules, bottle] \\
\midrule
\textbf{Result} & 
\textit{Success} (5 steps, Reward=1.00) & 
\textit{Partial Match} (Missed quantity, Reward=0.75) & 
\textit{Inefficient} (Multiple errors, Reward=0.25) \\
\midrule
 & \textbf{Ours} & \textbf{MoA} & \textbf{LATS} \\
\bottomrule
\end{tabular}
\caption{Case Study on a WebShop Task}  
\label{tab:case-study}
\end{table*}

% \begin{table*}[ht]
% \centering
% \setlength{\tabcolsep}{1.5pt} % 压缩列间距（合规操作）
% \begin{tabular}{p{0.12\textwidth} p{0.26\textwidth} p{0.26\textwidth} p{0.26\textwidth}}  % 缩减列宽
% \toprule
% \multicolumn{4}{c}{
% \parbox{0.9\textwidth}{ % 限制标题宽度，避免溢出
% \textbf{Instruction:} I'm looking for hair treatments that are sulfate and paraben free and are of high quality too. I need it in bottle form with 60 capsules, and price lower than \$70.00.
% }
% }\\
% \midrule

% \textbf{1. Query} & 
% search[sulfate and paraben free hair treatments 60 capsules bottle price \textless 70.00.] & 
% search[hair treatments sulfate paraben free 60 capsules bottle high quality price \textless 70.00] & 
% search[hair treatment sulfate free paraben free high quality 60 capsules] \\

% \midrule

% \textbf{2. Filter} & 
% click[Next \textgreater] & 
% click[B07MTPW3Z7] & 
% click[Back to Search] \\

% \midrule

% \textbf{3. Check} & 
% click[B09HWS67T8] & 
% click[1 bottle + 60 capsules] & 
% search[high quality hair treatment sulfate-free paraben-free 60 capsules] \\

% \midrule

% \textbf{4. Final} & 
% click[1 bottle + 60 capsules]... click[Buy Now] & 
% click[Buy Now] & 
% click[Back to Search]... search[hair treatment sulfate-free paraben-free 60 capsules bottle] \\

% \midrule

% \textbf{Result} & 
% \textit{Success} (5 steps, Reward=1.00) & 
% \textit{Partial Match} (Missed quantity, Reward=0.75) & 
% \textit{Inefficient} (Multiple errors, Reward=0.25) \\

% \midrule
%  & \textbf{Ours} & \textbf{MoA} & \textbf{LATS} \\

% \bottomrule
% \end{tabular}

% \caption{Case Study on a WebShop Task}  
% \label{tab:case-study}
% \end{table*}

To assess the contribution of key components in our proposed  framework, we conduct a series of ablation studies across two main dimensions: (i) the effectiveness of the Expert Council mechanism and its submodules, and (ii) the necessity of Dual-Signal MCTS in structured reasoning. All ablations are performed under controlled conditions on three benchmark datasets. %—Game of 24, WebShop, and HumanEval. We evaluate using success rate in the WebShop environment.

\textbf{Effectiveness of the Council Mechanism.}
To examine the effectiveness of our expert council mechanism, we conduct controlled ablation studies under a fixed dual-signal MCTS planner. % The analysis focuses on the following key components: council Size, routing strategies, and memory utilization. 
Results are reported in Table~\ref{tab:ablation_counsil}. Reducing the \textbf{council size} leads to marked performance drops, indicating that individual experts lack sufficient breadth or specialization to handle diverse task demands. Alternative \textbf{routing strategies}, such as random selection, round-robin scheduling, majority voting, and collaborative aggregation, fail to match the effectiveness of our task-aware routing, often misaligning expert capabilities with contextual requirements. Similarly, \textbf{memory utilization} strategies that exclude success segments or include failed trajectories significantly degrade performance, highlighting the importance of selective, high-quality experience reuse. These findings collectively affirm the importance of all three components. %  the full configuration, combining expert specialization, task-aligned routing, and targeted memory injection, achieves the most consistent and robust results.

\textbf{Effectiveness of Dual-signal MCTS.}
To assess the contribution of our dual-signal MCTS planner, we conduct ablation studies under a fixed council configuration. %The analysis focuses on the following key components: MCTS and dual-signal value estimation.  
Detailed results are shown in Table~\ref{tab:ablation_mcts}. Removing \textbf{MCTS} in favor of greedy, single-path selection results in substantial performance degradation, especially in tasks requiring long-horizon reasoning, due to the lack of lookahead and structural credit assignment. Similarly, replacing our \textbf{dual-signal value estimator} with single-signal variants—based solely on environment rewards, expert evaluation, or memory retrieval all leads to notable drops in accuracy, reflecting the limitations of sparse feedback or insufficient inductive grounding. These results underscore the necessity of both tree-based planning and hybrid value estimation. %  the full configuration consistently delivers superior performance by jointly leveraging structured search and complementary signals.

% We also tried constructing the council via temperature scaling and adversarial prompting from a shared base model. These yielded surface diversity but unstable reasoning and poorer performance. Detailed results are provided in the Appendix.\textcolor{red}{Appendix Number}
We also explored alternative council construction methods, such as temperature scaling and adversarial prompting from a shared base model, which led to superficial diversity but unstable reasoning and reduced performance. Details about council construction methods and hyperparameter tuning can be found in Appendix B and D.

\subsection{Efficiency Analysis}
% WebShop
% 这里可以用图，比如横坐标是树的深度，纵坐标是成功率
% baseline: LATS等树形方法，我们使用单信号的variant

% 这里需要考虑针对非树形搜索的方法，如何证明效率高？用token消耗？我们相对非树形的方法，token消耗占优吗？

\begin{figure}[ht] % 普通 figure 环境（非 figure*）
    \centering
    % 宽度设为当前栏宽的 0.8（按需调整，避免过宽）
    \includegraphics[width=0.77\columnwidth]{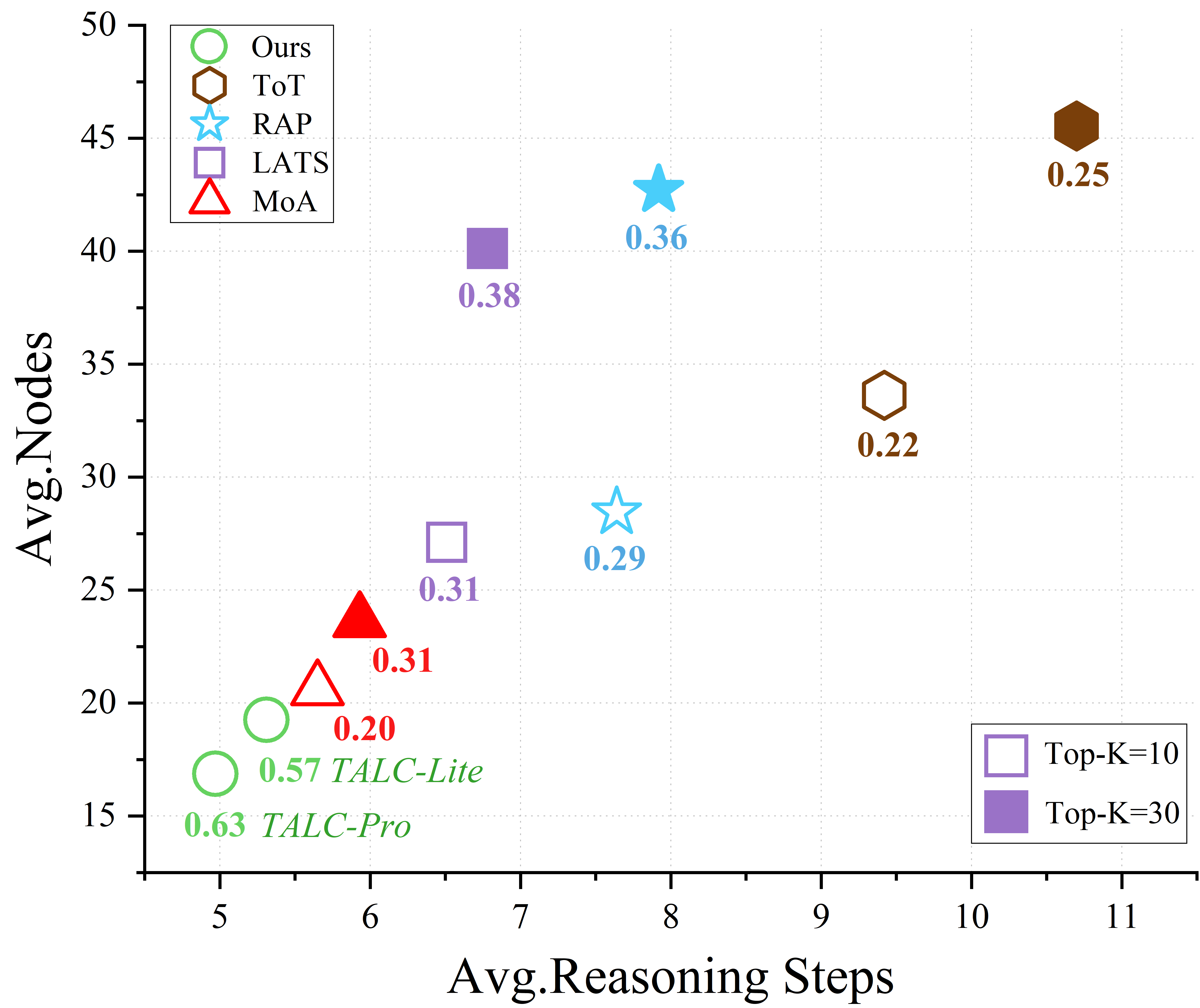}
    \caption{Efficiency Analysis. Baselines use K=10 and K=30 max iterations; our method uses only K=10.
}
    \label{fig:efficiency}
\end{figure}

We evaluate method efficiency on the WebShop task by comparing the maximum reasoning depth and total number of expanded nodes per successful case. For non-tree methods like MoA, each proposer-aggregator layer is counted as one step. For tree-based approaches (ToT, RAP, LATS, and TALC variants), we record the depth and total expanded nodes during search. Here, $K$ denotes the maximum number of trajectories allowed per task. While prior methods are run with a generous budget ($K=30$), our TALC-Pro variant uses only $K=10$ yet achieves the highest success rate ($0.63$) with the fewest reasoning steps ($4.97$) and nodes ($16.87$), outperforming all baselines in both effectiveness and efficiency. This highlights that our approach achieves strong planning with minimal budget, unlike other methods that trade accuracy for cost. We also provide a detailed comparison of token consumption in Appendix C.

% We evaluate the efficiency of different methods on the WebShop task by comparing the maximum number of reasoning steps and the total number of expanded nodes required to solve each task, based on successful case.
% For non-tree-based methods (MoA), we adopt its hierarchical planner structure and augment it with a sequential decision mechanism: each proposer-aggregator layer is treated as one reasoning step, and the total number of layers reflects the reasoning depth.
% For tree-based methods (ToT, RAP, LATS, and our TALC variants), we record the maximum depth of the search tree and the total number of nodes expanded during search.

% As shown in Figure~\ref{fig:efficiency}, TRLC-Pro achieves the best task success rate (0.63) while also requiring the fewest average reasoning steps (4.97) and the fewest expanded nodes (16.87). Notably, although other methods are allowed to use a higher inference budget ($K=30$), our method runs with $K=10$ and still outperforms all baselines in both effectiveness and efficiency. This highlights that our approach is not strongly dependent on large model capacity or high inference budget.

% We also observe that stronger planning methods (e.g., ToT, RAP, LATS) generally improve task accuracy but at the cost of significantly more steps and nodes. TALC, in contrast, achieves both better accuracy and lower computational overhead than even MoA, which lacks structured planning. This demonstrates that our trajectory-aware reasoning mechanism offers superior cost-effectiveness, especially in low-budget settings.

\subsection{Case Study}

We examine a representative task from the WebShop benchmark. This task involves multi-faceted constraints spanning product ingredients, packaging form, dosage count, and price cap. We compare three decision-making strategies: our full framework, a MoA-style variant with greedy single-step inference, and LATS, a non-specialized, single-model baseline. As shown in Table~\ref{tab:case-study}, only our method consistently satisfies all task constraints by leveraging expert specialization and tree-based planning to progressively narrow the candidate space and recover from early errors. The MoA-style agent, despite using expert routing, lacks trajectory-level coordination and often overlooks subtle constraints due to shallow reasoning. LATS performs worst, frequently issuing redundant or misaligned actions, failing to reconcile constraints over time. These results highlight the necessity of integrating structured planning with expert-driven decision pathways for precise, constraint-faithful reasoning.

\section{Conclusion}
% The preparation of the \LaTeX{} and Bib\TeX{} files that implement these instructions was supported by Schlumberger Palo Alto Research, AT\&T Bell Laboratories, Morgan

This paper proposes TALC, a task-aware decision framework that integrates expert routing with dual-signal planning. By leveraging a council of specialized LLMs and memory-guided value estimation, TALC enables adaptive, efficient reasoning across tasks. Experiments show that it consistently outperforms strong baselines under both lightweight and full configurations, demonstrating the effectiveness of expert coordination and structured search.

\newpage
\bibliography{aaai2026} % 指定AAAI样式

% \newpage
% \appendix
% \input{appendix}

\end{document}